\documentclass[10pt]{cai}
\usepackage{hyperref}
\hypersetup{
    colorlinks=true,
    linkcolor=magenta,
    filecolor=red,      
    urlcolor=black,
    pdftitle={Overleaf Example},
    pdfpagemode=FullScreen,
    }

\begin{document}
\def\conferenceyear{2025}
\volumeheader{38}{}
\begin{center}

\title{Exploring Superposition and Interference in State-of-the-Art Low-Parameter Vision Models}
\maketitle

\thispagestyle{empty}

\begin{tabular}{cc}
Lilian Hollard\upstairs{\affilone,*}, Lucas Mohimont\upstairs{\affilone}, Nathalie Gaveau\upstairs{\affiltwo}, Luiz-Angelo Steffenel\upstairs{\affilone}
\\[0.25ex]
{\small \upstairs{\affilone} Université de Reims Champagne-Ardenne, CEA, LRC DIGIT, LICIIS, Reims, France } \\
{\small \upstairs{\affiltwo} Université de Reims Champagne-Ardenne, INRAE, RIBP USC 1488, Reims, France} \\
\end{tabular}
  
\emails{
  \upstairs{*}lilian.hollard@univ-reims.fr 
}
\vspace*{0.2in}
\end{center}

\begin{abstract}
The paper investigates the performance of state-of-the-art low-parameter deep neural networks for computer vision, focusing on bottleneck architectures and their behavior using superlinear activation functions. We address interference in feature maps, a phenomenon associated with superposition, where neurons simultaneously encode multiple characteristics. Our research suggests that limiting interference can enhance scaling and accuracy in very low-scaled networks (under 1.5M parameters). We identify key design elements that reduce interference by examining various bottleneck architectures, leading to a more efficient neural network. Consequently, we propose a proof-of-concept architecture named NoDepth Bottleneck built on mechanistic insights from our experiments, demonstrating robust scaling accuracy on the ImageNet dataset. These findings contribute to more efficient and scalable neural networks for the low-parameter range and advance the understanding of bottlenecks in computer vision. \url{ https://caiac.pubpub.org/pub/3dh6rsel}

\end{abstract}

\copyrightnotice

\section{Introduction}
Understanding the mechanistic behavior of deep neural networks, especially in high-dimensional spaces, remains a significant challenge. Researchers need precise knowledge of how many parameters, filters, and weights are needed to achieve a specific level of accuracy. This uncertainty is evident in computer vision, where the image data is naturally unpredictable, making it difficult to determine the precise amount of information required to remove features. If we understand how different neural network architectures behave, we can optimize and reduce the required parameters for a given task.

Reducing parameters is critical in areas like robotic vision and Edge AI, where reducing computational costs is essential to enable real-time inference on low-power devices. While Edge AI offers advantages over cloud-based solutions—improved security, reduced latency, and lower data transfer costs \cite{buyya_cloud_2009}—deploying deep learning models on resource-constrained hardware requires efficient neural architectures. This need has led to the rise of mobile-friendly neural networks, which use fewer parameters and computational optimizations to maintain performance.

Instead of relying solely on expensive Neural Architecture Search (NAS) techniques to reduce parameters, an emerging direction involves mechanistic interpretability\footnote{Focus on discovering the underlying mechanisms and algorithms that shape the network's behavior.}—specifically, the concept of superposition \cite{elhage2022toy}. Superposition describes how neural networks can represent more features than their available dimensions allow, leading to polysemantic neurons—where a single neuron encodes multiple features. 

Understanding superposition in complex tasks like CIFAR \cite{Krizhevsky2009LearningML} or ImageNet \cite{deng_imagenet_2009} classification remains nearly impossible, especially with deep neural networks. Therefore, instead of observing how superposition occurs, we propose a reverse approach: \textit{we study the behavior of neural network architecture when we attempt to prevent superposition.} 

Our work explores the mechanistic interpretability of bottleneck architectures (Inverted, Sandglass, and ConvNeXt-like Bottlenecks) in computer vision. We aim to reverse-engineer the core mechanisms of networks built on bottleneck structures by utilizing superlinear activation functions. Superlinear activation functions amplify neuron activations at a rate greater than linear growth as pre-activation values increase. A notable example is the SoLU module (Softmax Linear Unit) \cite{elhage2022solu}, which enables us to analyze how low-cost neural networks depend on interferences to facilitate superposition. We examine the impact of SoLU on network behavior and performance by constraining interference.
In our experiments, we identify key architectural design elements that minimize interference, removing the need for SoLU. In doing so, we provide several intuitions about effective shallow-scaling neural networks. Our findings align closely with observations from state-of-the-art low-parameter computer vision classifiers. To further support our observations, we introduce a proof-of-concept called the NoDepth Bottleneck, which empirically validates our results on ImageNet.

Our key contributions are the following:
\begin{enumerate}
    \item \textbf{Interference Dependence:} We found that MobileNeXt (Batch Norm), Mobile\-Netv2, and MobileNetv3 scale poorly due to their reliance on interference, which is ineffective at low parameters.
    \item \textbf{Neural Network Architecture:} Reducing interference using interference-free architecture improved scaling, especially at low parameters. Models like ConvNeXt, MobileNeXt (Layer Norm), and NoDepth Bottleneck showed minimal impact.
    \item \textbf{Feature Alignment:} Feature map analysis revealed that models most affected by SoLU activated multiple features in non-natural high-dimensional spaces compared to their base implementation. SoLU's noise reduction confirmed their reliance on interference alongside stronger features.
\end{enumerate}

This paper is structured as follows: Section \ref{sec:literature} explores related work, defining superposition and methods to mitigate it. Section \ref{sec:exp} introduces the bottleneck architectures central to our study and explains their relevance. In Section \ref{sec:exp_results}, we conduct initial observations on accuracy variations when preventing superposition in the activation space within the residual stream. Section \ref{sec:exp_results_depth} examines the presence of quasi-orthogonal vectors, identifying architectures that naturally align with superlinear activation functions without additional activations. We compare Batch Norm and Layer Norm, highlighting why Layer Norm is more stable. In Section \ref{sec:exp_results_aligned}, we demonstrate that neural network aligns their features in a particular behavior (orthogonal or not)—their dependence on interference drastically changes the conduct of their feature alignment. Finally, Section \ref{sec:proofofconcept} presents a proof of concept on a larger-scale dataset to validate our findings, followed by a discussion on potential future directions in Section \ref{sec:conclusion}.

\section{Literature Review}
\label{sec:literature}
Utilizing superlinear activation functions—specifically the SoLU activation—to minimize feature map interference and analyze the mechanistic efficiency of different bottleneck architectures presents a novel perspective. 
We first review prior work on low-parameter neural network architectures and their reliance on bottlenecks in computer vision. Finally, we reference existing definitions of superposition, interference, and the SoLU module to establish a foundation.

\subsection{State-of-the-Art Deep Neural Network Architecture} \label{sec:literature_sota}

Deep Learning's automatic feature learning is crucial with the explosion of visual data from social media \cite{monti2019fake, lazreg2016deep}, surveillance \cite{chen2017deep, wu2018real}, and medical imaging \cite{suzuki2017overview, anaya2021overview}. Its scalability enhances applications in security, entertainment, agriculture \cite{mohimont2023accurate, hollard2023applying}, and art \cite{wang2017generative, goodfellow2020generative}, where accurate image analysis is essential.
Two key innovations—convolutional operations \cite{lecun1995convolutional} and self-attention mechanisms \cite{vaswani_attention_2017}—transformed computer vision. However, deep networks remain computationally demanding, limiting the deployment of Transformer architecture on low-power devices. While cloud computing \cite{buyya_cloud_2009} and fog computing \cite{steffenel_improving_2018} help using deeper neural networks, they introduce latency, bandwidth issues, and security risks \cite{buyya_cloud_2009, zhou_edge_2019, wang_deep_2020}.
To address this, optimized AI architectures reduce cloud reliance, improving security, lowering latency, and cutting power consumption \cite{singh2023edge, gill2025edge}. Efficient vision models are essential for embedded devices, which must process data locally despite strict resource constraints.

\textbf{Advancements in Low-Parameter Neural Network Models.} From MobileNetv1 \cite{mehta_mobilevit_2022} to MobileViTv3 \cite{wadekar_mobilevitv3_2022}, numerous low-cost neural network architectures have emerged, including MobileNets \cite{howard_mobilenets_2017, sandler_mobilenetv2_2018,howard_searching_2019}, EfficientNets \cite{tan_efficientnet_2019, tan_efficientnetv2_2021}, and MobileViT \cite{mehta_mobilevit_2022, mehta_separable_2022,wadekar_mobilevitv3_2022}, among others. 
Since the introduction of early pioneers in optimizing deep neural networks, researchers have continuously worked to achieve high performance on complex image datasets such as ImageNet while reducing computational costs.
MobileNetv1 introduced depthwise separable convolutions, drastically reducing computational costs. This trend accelerated between 2020 and 2023 with models like MobileViT \cite{mehta_mobilevit_2022}, which fused Transformer-based mechanisms with MobileNet principles for better efficiency and accuracy. MobileViT achieves 80\% Top-1 accuracy on ImageNet with just 5.1M parameters, while ViT needs over 80M to reach 77\%. These advancements influenced large-scale models like ConvNeXt \cite{liu_convnet_2022}, integrating parameter-efficient techniques while maintaining high performance. Compact AI extends beyond edge devices, offering scalable and efficient solutions across AI research and deployment.

\begin{table}[htbp]
    \centering
    \begin{minipage}{0.45\textwidth}
        \centering
        \begin{tabular}{|l|c|c|}
            \hline
            Models&Parameters(M)&Top-1 Accuracy\\
            \hline
            MobileNetv1& 0.5&50.6\\
            \hline
            MobileNetv1 & 1.3&63.7\\
            MobileViT & 1.3&69\\
            MobileNetv3  & 1.6 &58\\
            MobileNetv2 & 1.7&60.3\\
            MobileNeXt & 1.8&64.7\\
            \hline
            MobileNetv2&2.2&66.7\\
            ShuffleNetv2& 2.4&66.2\\
            MobileNeXt & 2.5&72\\
            MobileNetv3 & 2.5 &67.5\\
            MobileNetv1 & 2.6 & 68.4\\
            GhostNet & 2.6 & 66.2\\
            \hline
            MobileNetv3 & 3.6&70.4\\
            EfficientNetB0 &5.2&77.692\\
            \hline
        \end{tabular}
        \caption{Complete State-of-the-art of low-parameter networks}
        \label{tab:completesota}
    \end{minipage}
    \hfill
    \begin{minipage}{0.48\textwidth}
        \centering
        \includegraphics[width=1\linewidth]{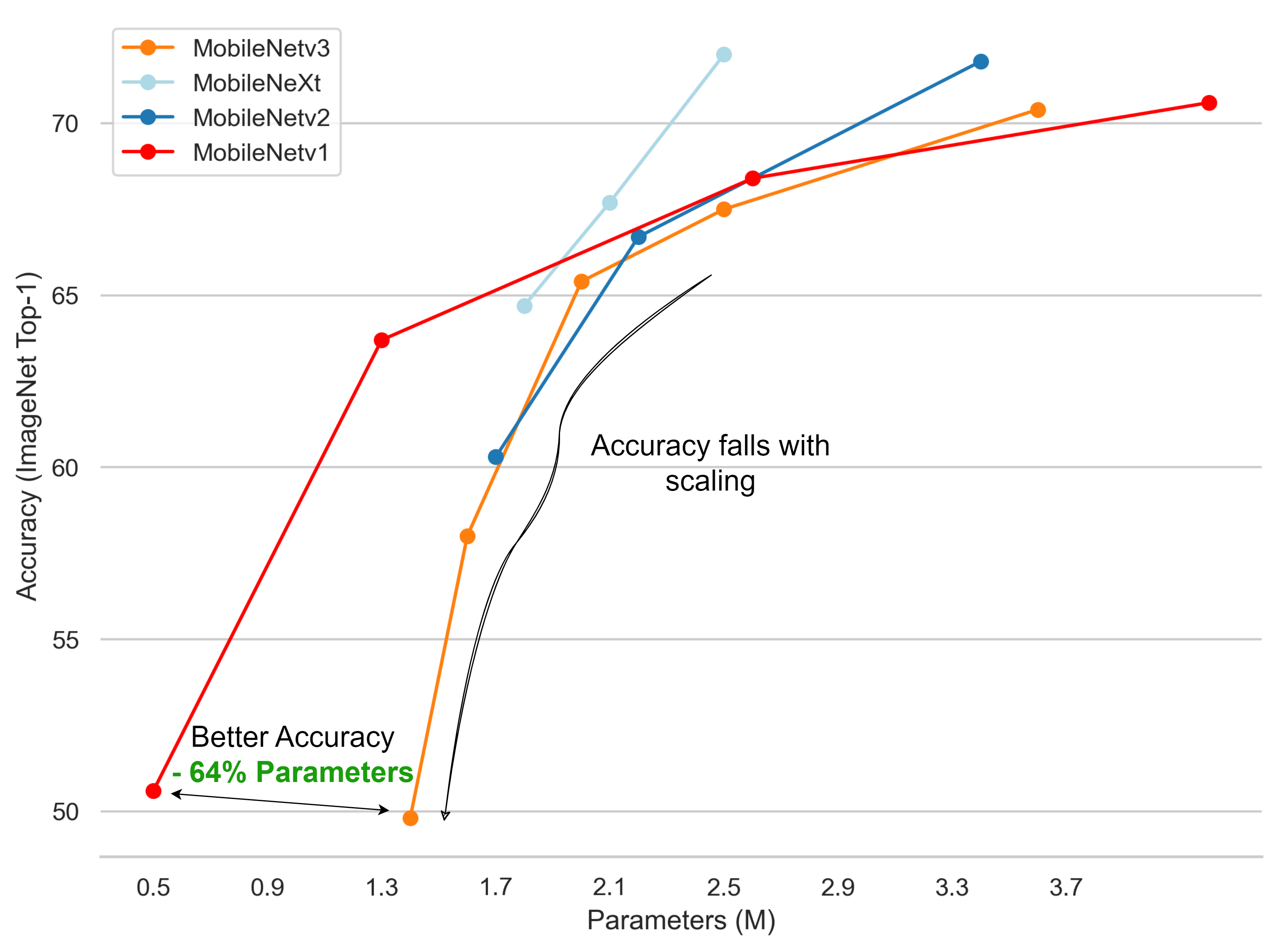}
        \captionof{figure}{Inverted Bottleneck scaling issue}
        \label{fig:IBIssue}
    \end{minipage}
\end{table}

When examining ultra-low-parameter neural networks (ranging from 0.5M to 1.5M parameters), an important pattern is often overlooked: scaling. It is commonly assumed that MobileNetv2 and v3 progressively improve the parameter-to-accuracy ratio. While this holds true at larger scales, MobileNetv1 outperforms both at lower parameter counts. For instance, with 1.5M parameters, MobileNetv1 achieves \textbf{13.9 points} higher accuracy than MobileNetv3 (Table \ref{tab:completesota} - Figure \ref{fig:IBIssue}).  

This explains why many edge devices and microcontrollers continue to use MobileNetv1 as a benchmark, despite its architecture being nearly \textit{eight years old}. A similar trend appears with MobileNeXt \cite{dequan_rethinking}, which introduces the sandglass module to refine the Inverted Bottleneck. While it scales well at higher parameter counts (relatively to other low-parameter networks), it still falls short compared to MobileNetv1 (Figure \ref{fig:IBIssue}).

\textbf{Features alignment study.}
Liu et al. \cite{liu2023efficientvit} and Zhang et al. \cite{zhang2021sparse} both explore improving Transformer efficiency but from different angles. Liu et al. \cite{liu2023efficientvit} reduces cosine similarity within Query, Key, and Value vectors by modifying the architecture beyond a standard MLP, enhancing accuracy without investigating the underlying mechanisms. Meanwhile, Zhang et al. \cite{zhang2021sparse} introduce a sparser attention module for NLP, showing that increased sparsity improves network comprehension and precision. However, their method relies on ReLU and Layer Norm rather than superlinear activation functions and focuses on refining the softmax function rather than understanding architectural behavior.

\subsection{Prior Definitions}
\begin{figure}[ht]
    \centering
    \includegraphics[width=1\linewidth]{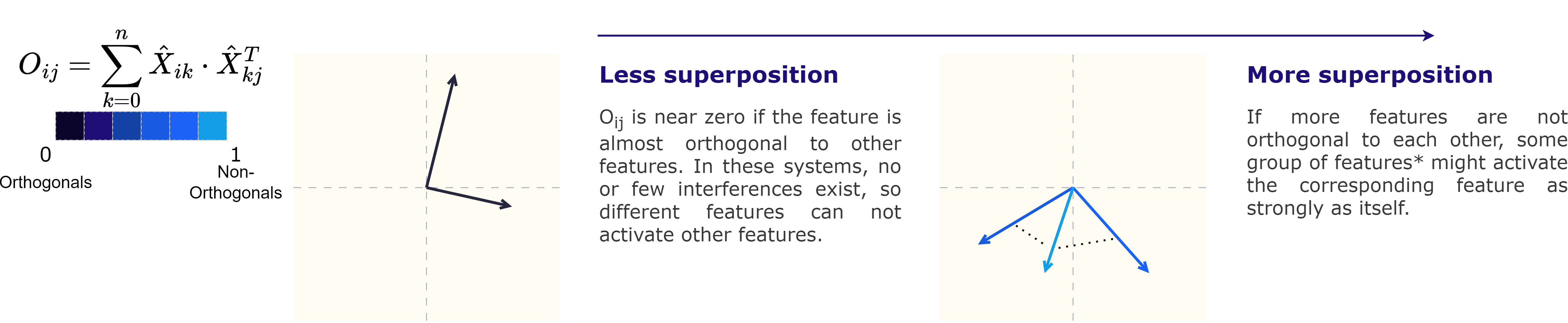}
    \caption{2D visualization of feature alignment and interference: \( O_{ij} \) ranges from 0 for orthogonal features and 1 for high superposition, leading to unintended activations.\\ *Interferences}
    \label{fig:interference}
\end{figure}
\textbf{Superposition and Interferences.} In neural networks, superposition occurs when a single neuron or group of neurons simultaneously represent multiple distinct features, a phenomenon known as polysemanticity—where one neuron encodes multiple types of information \cite{elhage2022toy}. A neural network encodes overlapping directions within a high-dimensional space using non-orthogonal features (represented as 2d vectors in Figure \ref{fig:interference}), which are not strictly orthogonal and use the same dimensional space. The difference between these features packed in the same dimension is interference. 
Quasi-orthogonal vectors, where pairwise dot products are close to zero, result in a weak superposition state, as the network underutilizes interference. The more orthogonal the features, the less they interfere, leading to minimal overlap in representations.

From this point forward, we define interference as a value ranging from 0 to 1, where 0 indicates a feature that does not interfere with another, and 1 represents a feature that is fully aligned with another in the same high-dimensional space. Interference increases if the network packs more features into the same dimensional space.

\textbf{Softmax Linear Unit (SoLU).} We seek to identify architectural designs that harness the superposition phenomenon to better understand the inner workings of bottlenecks. Our primary approach is to limit the bottleneck's reliance on interference without entirely eliminating it. Completely suppressing interference would reduce performance to that of linear networks \cite{saxe2013exact}, which is counterproductive. To mitigate interference while preserving useful feature representations, we employ the Softmax Linear Unit (SoLU) \cite{elhage2022solu}—a superlinear activation function that dynamically adjusts the activation space to minimize unwanted overlaps. Originally introduced to reduce superposition in Transformer models for interpretability, we extend SoLU's application to convolutional bottlenecks. This key step could open new avenues for research on how different architectures respond to superlinear activations regarding efficiency, interpretability, and sparsity.

\begin{figure}[ht]
    \centering
    \includegraphics[width=0.75\linewidth]{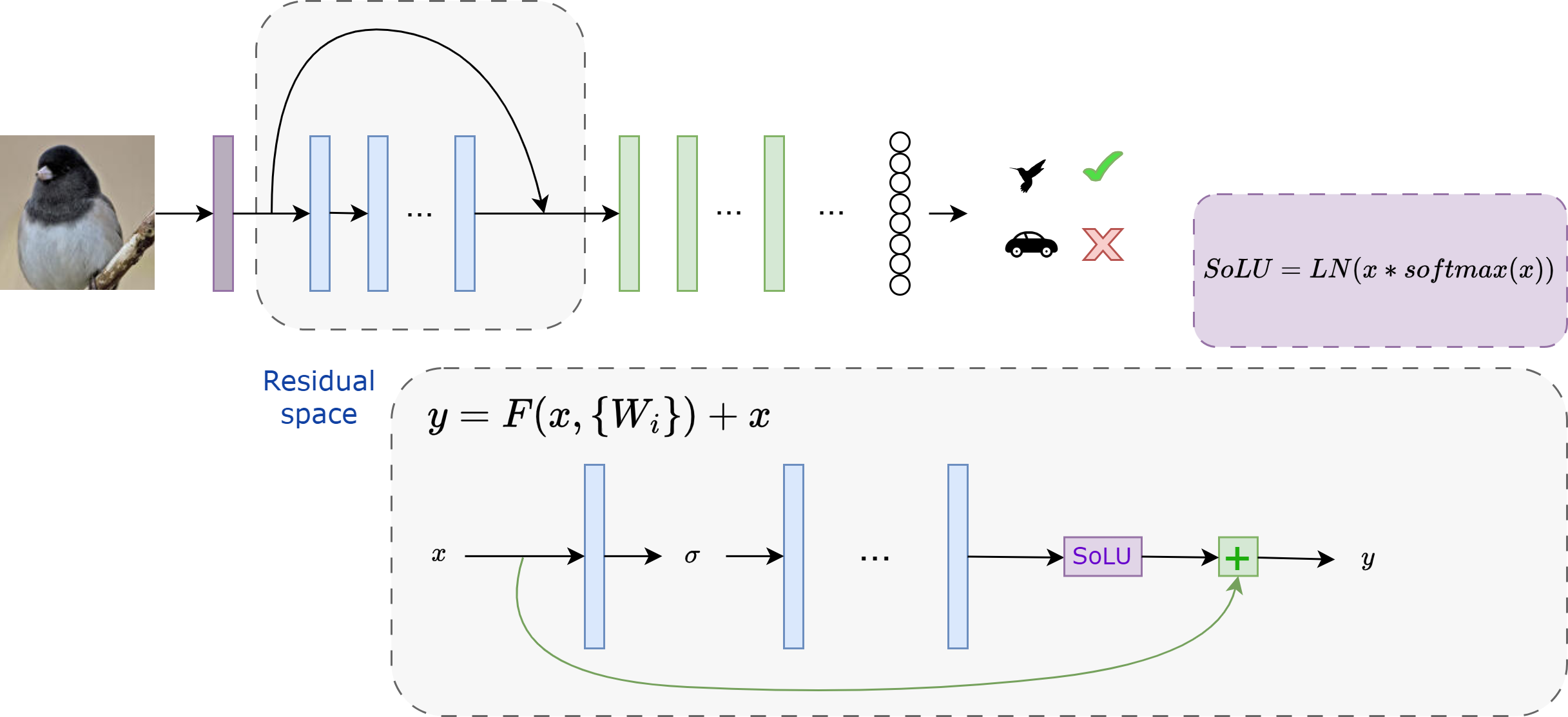}
    \caption{Our proposition of the SoLU module integration into residual space}
    \label{fig:ressolu}
\end{figure}

We propose integrating the SoLU function\footnote{In equations \ref{eq:lnsolu} and \ref{eq:lnsolu_residual}, $ln$ denotes Layer Norm.}, as described in equation \ref{eq:lnsolu}, into various low-parameter state-of-the-art architectures (Figure \ref{fig:ressolu}). 
\begin{equation*}
    SoLU = x * softmax(x)
\end{equation*}
\begin{equation}\label{eq:lnsolu}
    f(x) = ln(SoLU(x)) = ln(x * softmax(x))
\end{equation}

Specifically, we incorporate SoLU into the residual stream of each bottleneck architecture $y=F(x, \{W_i\})+x$ just before the residual connection as equation \ref{eq:lnsolu_residual}. 

\begin{equation}\label{eq:lnsolu_residual}
    y=ln(SoLU(F(x, \{W_i\})))+x 
\end{equation}

\section{Experimental Method}
\label{sec:exp}
Our experiments focus on bottlenecks, which are the result of dimensional expansions or reductions using convolutions. In state-of-the-art low-cost neural networks presented earlier, bottlenecks play a critical role in optimizing parameter counts.

We aim to study the impact of the SoLU module on three types of bottlenecks: Inverted Bottlenecks (MobileNetv2 and v3) \cite{sandler_mobilenetv2_2018, wadekar_mobilevitv3_2022}, Sandglass Bottlenecks \cite{dequan_rethinking}, and ConvNeXt-like Bottlenecks \cite{liu_convnet_2022}. 

\textbf{Inverted Bottleneck:} Introduced by MobileNetv2 and optimized by MobileNetv3, the Inverted Bottleneck is an effective method for creating high-accuracy models with low parameter costs. It utilizes depthwise convolution ($W_{dw} \in \mathbb{R}^{k,k,1,C \times \textcolor{magenta}{\alpha}}$), which computes convolution with a large kernel size to capture spatial information but only processes one channel at a time (instead of the entire channel dimension as in standard convolutions). Two pointwise convolutions ($W_{in} \in \mathbb{R}^{1,1,C \times \textcolor{magenta}{\alpha}, C}$ and $W_{out} \in \mathbb{R}^{1,1,C, C \times \textcolor{magenta}{\alpha}}$) manage channel dimensions by expanding or reducing the channels with an expansion ratio ($\textcolor{magenta}{\alpha} \geq 2.0 $). Additionally, all convolutions are followed by Batch Norm ($bn$), as seen in both MobileNetv2 and MobileNetv3.
\begin{equation}\label{eq:inverted_bottleneck}
    y = F(x, \{W_i\})+x=bn( W_{out}(\sigma(bn(W_{dw}(\sigma(bn(W_{in}(x)))))))) + x
\end{equation}
\newline    
\textbf{ConvNeXt-like Bottleneck:} This design separates the depthwise convolution $W_{dw} \in \mathbb{R}^{k,k,1,C}$ from the pointwise convolutions used for expansion $W_{in} \in \mathbb{R}^{1,1,C \times \alpha, C}$ and reduction $W_{in} \in \mathbb{R}^{1,1,C,C \times \alpha}$. The approach ensures a clear distinction between operations while maintaining efficiency and flexibility. 

Additionally, ConvNeXt uses fewer activation functions ($\sigma$) and replaces Batch Norm with Layer Norm, a change we will examine to determine its impact on superposition.
\begin{equation}
    y = F(x, \{W_i\})+x = W_{out}(\sigma(W_{in}(ln(W_{dw}(x)))) + x
\end{equation}
\textbf{Sandglass Bottleneck: } Introduced in MobileNeXt, depthwise convolutions operate in a high-dimensional input space ($C$), not in one created by a 1x1 expansion (as $\alpha \leq 1.0$, leading to a lower-dimensional space within pointwise convolutions). According to the authors, this approach also improves accuracy. It is interesting to contrast this with ConvNeXt, where the depthwise convolution also matches the dimension at the start of the residual stream. However, in ConvNeXt, this occurs in a lower-dimensional space than the expanded dimensions created by the expansion ratio.
\begin{equation}
    y = F(x, \{W_i\})+x = bn\textcolor{orange}{(}W_{dw_2}(bn\textcolor{magenta}{(}W_{out}(\sigma(bn\textcolor{red}{(}W_{in}(bn\textcolor{blue}{(}W_{dw_1}(x)\textcolor{blue}{)})\textcolor{red}{)}))\textcolor{magenta}{)})\textcolor{orange}{)} + x
\end{equation}

\subsubsection{Models vessel}
\begin{figure}[ht]
    \centering
    \includegraphics[width=1.0\linewidth]{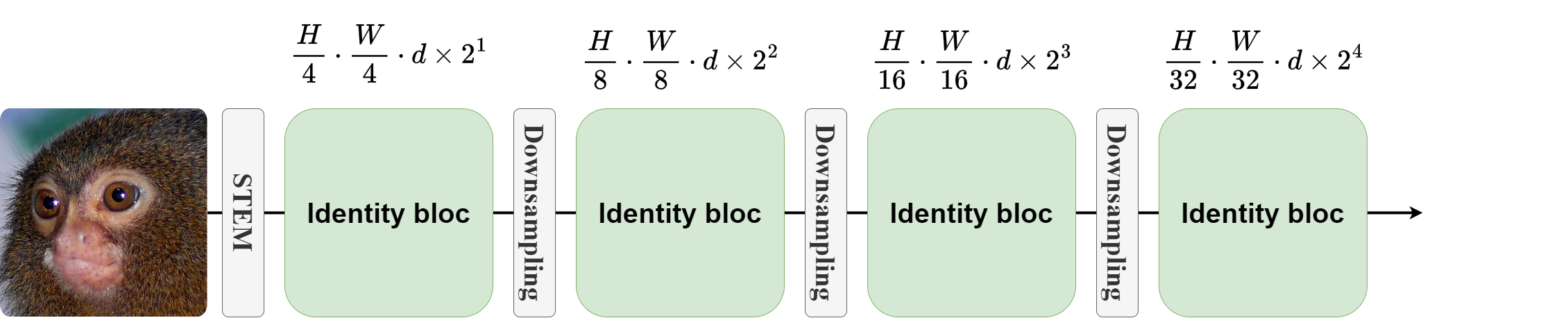}
    \caption{The base vessel containing each specific architecture, inspired by ConvNeXt \cite{liu_convnet_2022}. We use a depth repetition for each identity bloc of [3,3,9,3].}
    \label{fig:vessel}
\end{figure}
Directly comparing different bottlenecks using their base models would be ineffective, as each architecture has unique characteristics, such as downsampling methods, initial layer configurations, and varying channel counts or depth repetitions.
Instead, we use a unified framework integrating different bottleneck structures to address the previous issue. Built upon ConvNeXT—which incorporates best practices from Vision Transformers—our experimental setup, shown in Figure \ref{fig:vessel}, serves as a standardized vessel for evaluating each bottleneck. We implement rapid spatial reduction using standard convolutions in the early layers and downsampling modules between identity blocks. These operations remain separate from the "identity blocks"—the architectural bottlenecks defined earlier—ensuring a fair and isolated comparison.

Our experiments using CIFAR-10, CIFAR-100, and ImageNet consistently demonstrated the same behavior. For the sake of computational efficiency and rapid experimentation, we use CIFAR-10 as our primary benchmark.
For all models and training on CIFAR-10, we maintained a consistent setup aligned with ImageNet and modern neural networks: a learning rate of 1e-4 with Adam and a cosine learning rate scheduler. While fine-tuning hyperparameters could enhance individual network performance, our objective is to analyze the mechanisms of bottlenecks rather than maximize accuracy. Varying learning rates or optimizers do not alter our core findings.

\section{Results}
\label{sec:exp_results}

\subsection{Depthwise convolution should stay in the input dimensional space}
\label{sec:exp_results_depth}

\begin{figure}[ht]
    \centering
    \includegraphics[width=1\linewidth]{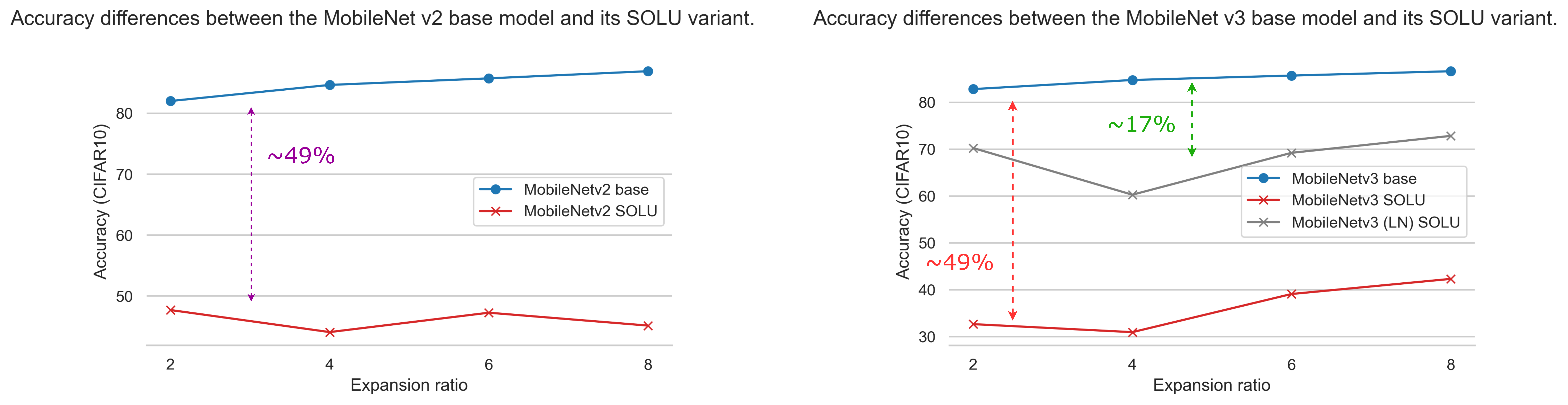}
    \caption{Accuracy difference between the base architecture of the Inverted Bottlenecks (v2 and v3) and their SoLU implementation.}
    \label{fig:mbv2_v3}
\end{figure}

Using an Inverted Bottleneck (whether v2 or v3, regardless of the activation function $\sigma$), we find that model accuracy is highly sensitive to interferences. Figure \ref{fig:mbv2_v3} shows that when we reduce or eliminate interference using the SoLU module, accuracy drops by nearly 50\% compared to the baseline approach.

\begin{figure}[ht]
    \centering
    \includegraphics[width=1\linewidth]{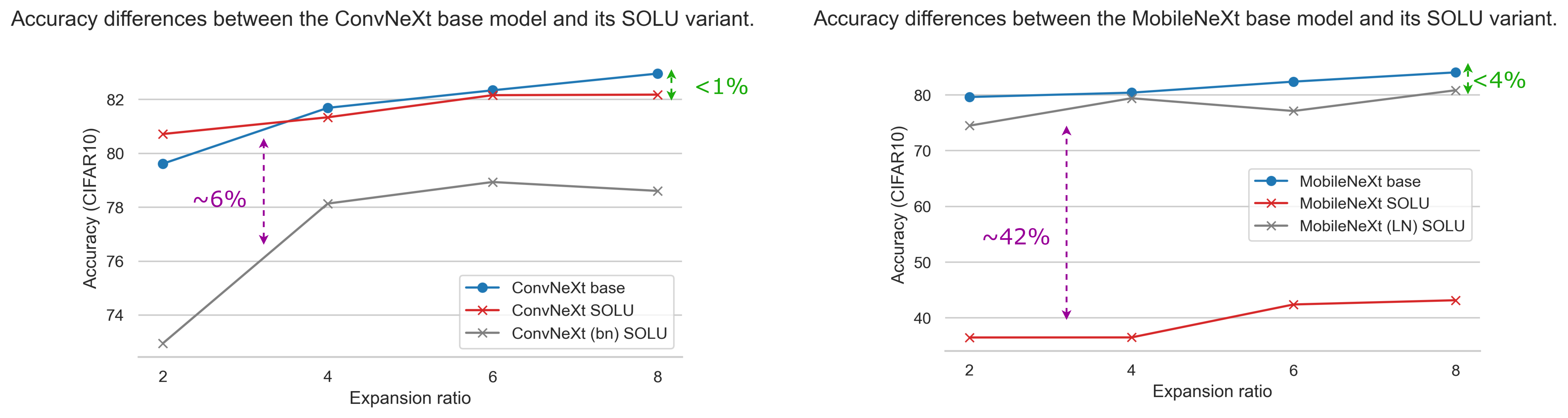}
    \caption{Accuracy difference between the base architecture of the ConvNeXt-like Bottleneck and Sandglass Bottleneck and their SoLU implementation.}
    \label{fig:cnext_and_mbnext}
\end{figure}

However, when analyzing ConvNeXt and MobileNeXt (with Layer Norm) in Figure \ref{fig:cnext_and_mbnext}, we observe that interference reduction has no effect when depthwise convolution operates at the same channel dimension as the input in the residual stream ($x \in \mathbb{R}^{B,\textcolor{red}{C},H,W}$, $W_{dw}\in \mathbb{R}^{k,k,1,\textcolor{red}{C}}$).

To explore whether the base architecture inherently mitigates interference, we simulate several configurations: a standard bottleneck with ReLU and Batch Norm within ConvNeXt, as well as Inverted and Sandglass Bottlenecks using Layer Norm in place of the traditional Batch Norm.
Various design considerations influence the choice between Batch Norm and Layer Norm. Notably, recent trends in compact vision architectures—such as ConvNeXt and its variants—have increasingly adopted Layer Norm, even for image data \cite{mehta_mobilevit_2022, liu2023efficientvit, graham_levit_2021}.
Across multiple scaling scenarios, ConvNeXt exhibits only a 6\% average accuracy drop despite Batch Norm’s known instability \cite{zhang2021sparse}. Interestingly, substituting Layer Norm into the Inverted Bottleneck—similar to ConvNeXt's approach—results in a significantly larger accuracy drop of 17\%. This suggests that Layer Norm alone is not responsible for ConvNeXt’s strong performance with SoLU; the architecture plays the key role.
We reinforce this insight with the Sandglass Bottleneck experiment, which shows only a 4\% accuracy loss. Overall, our experiments reveal that Batch Norm amplifies interference across architectures, especially when applied after high-dimensional pointwise expansion and reduction—where we observe an accuracy drop as severe as 42\%.

\subsection{Preventing superposition significantly disrupts neural network representational structure.}
\label{sec:exp_results_aligned}

We analyze feature representations in high-dimensional space to explore each layer of the chosen architecture, with or without SoLU. Each feature has a corresponding representation direction $X_i$ along the channel axis in the image representation. The model enters a superposition state when multiple features, such as $X_1$ and $X_2$, activate with similar values.

It is essential to clarify the notation: we are not referring to the direction of the weight $W$ but rather the feature maps $X$ produced by any $W$ on input $x$.
To determine if a feature shares its dimension with others, we compute $O_{ij}$ (eq. \ref{eq:oij}). If $O_{ij} = 0$, the feature is orthogonal to another. If $O_{ij} = 1$, they align in the same high-dimensional space (with $X$ normalized into $\hat{X}=\sqrt{\sum^{n}_{k=0}|X_k|^2}$). Values closer to 1 indicate that other features activate $X_i$ at a similar strength.
\begin{equation} \label{eq:oij}
    O_{ij}=\sum^{n}_{k=0}\hat{X}_{ik} \cdot \hat{X}^{T}_{kj}
\end{equation}
We examine the dot product between $X_i$ and other representation directions $X_j$ in both the base architecture and the one using SoLU to analyze how the feature alignment representation behaves when interference is constrained.

\begin{figure}[ht]
    \centering
    \includegraphics[width=1\linewidth]{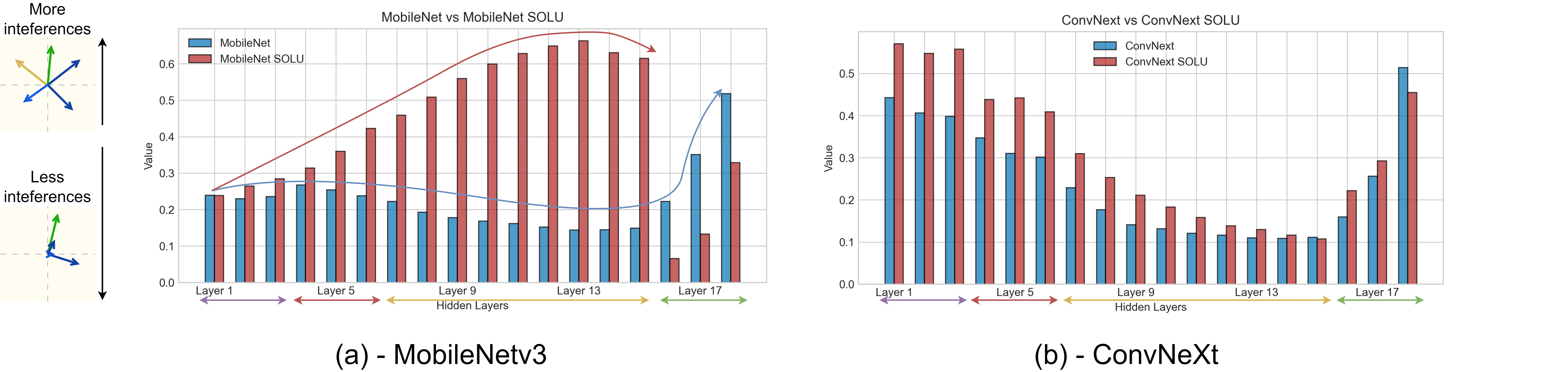}
    \caption{Mean features angles between architecture dependent on small interference (MobileNetv3) and interference-free architecture (ConvNeXt).}
    \label{fig:mbv3_et_next_direction}
\end{figure}

\textbf{MobileNetv3} exhibits a 50\% accuracy drop in its baseline form with Batch Norm. The model’s feature maps consist mostly of orthogonal vectors across the channel dimension. As shown in Figure \ref{fig:mbv3_et_next_direction}, the network attempts to push features into distinct directions within high-dimensional space, with $O_{ij}$ values ranging between 0 and 0.5—indicating that the network encodes independent features.
However, the network aligns features more closely when we train the same architecture with SoLU to suppress interference noise. This aligns with the known behavior of SoLU in the presence of polysemantic neurons. For instance, if neuron A is responsible for both feature “a” and feature “b,” then the combined input becomes $a+b$ during co-occurrence \cite{elhage2022solu}. Because SoLU is superadditive—i.e., $f(a+b)>f(a)+f(b)$—the neuron responds disproportionately, effectively "overreacting." Feature direction analysis under SoLU confirms that the model starts to rely on interference, as it must fire more strongly in a single direction. This makes it harder for the network to disentangle co-occurring features cleanly.

\textbf{ConvNeXt} provide valuable insights. Most features organize themselves in an orthogonal manner. Since the SoLU module has no impact (1\%) on ConvNeXt accuracies, we assume that in the feature map space, slight noise or interference naturally already aligns within their own direction space. Figure \ref{fig:mbv3_et_next_direction} illustrates that the network strives to distribute features across distinct spaces and directions, following the same behavior as the base ConvNeXt.

\begin{figure}
    \centering
    \includegraphics[width=1\linewidth]{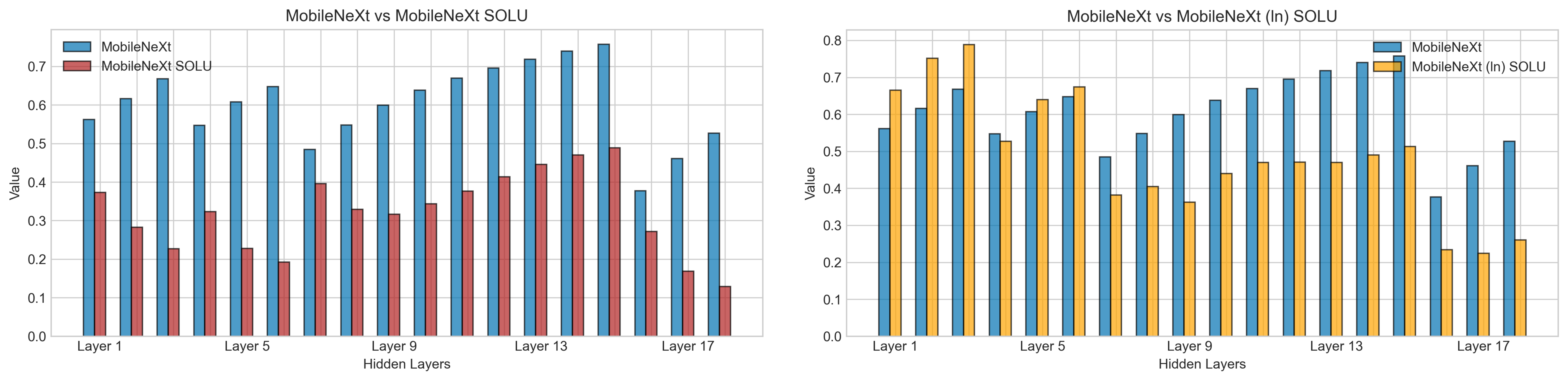}
    \caption{MobileNeXt features direction specificities.}
    \label{fig:mbnextortho}
\end{figure}

\textbf{MobileNeXt} follows a similar pattern: the base model, which uses Batch Norm, performs poorly when paired with the SoLU module. As illustrated in Figure \ref{fig:mbnextortho}, MobileNeXt utilizes high-dimensional space by aligning features in similar directions—over 50\% of layers exhibit this behavior. However, SoLU disrupts the architecture’s reliance on such interference-based representations. 
Initially, we hypothesized that SoLU would consistently encourage feature alignment in the same direction. Instead, we find that SoLU interferes with architectures that depend on interference-driven representations—effectively pushing features into suboptimal spaces for the given design.
To further support this observation, we replaced Batch Norm with Layer Norm in MobileNeXt. This change resulted in only a 4\% average accuracy drop compared to the 42\% drop with Batch Norm. In this system, the network maintained a similar high-dimensional feature organization as in the base model, and SoLU did not significantly disrupt its representational structure.

\subsection{Proof of concept: NoDepth Bottleneck}
\label{sec:proofofconcept}


\begin{figure}[ht]
\centering
\begin{minipage}{.49\textwidth}
  \centering
  \includegraphics[width=1.0\linewidth]{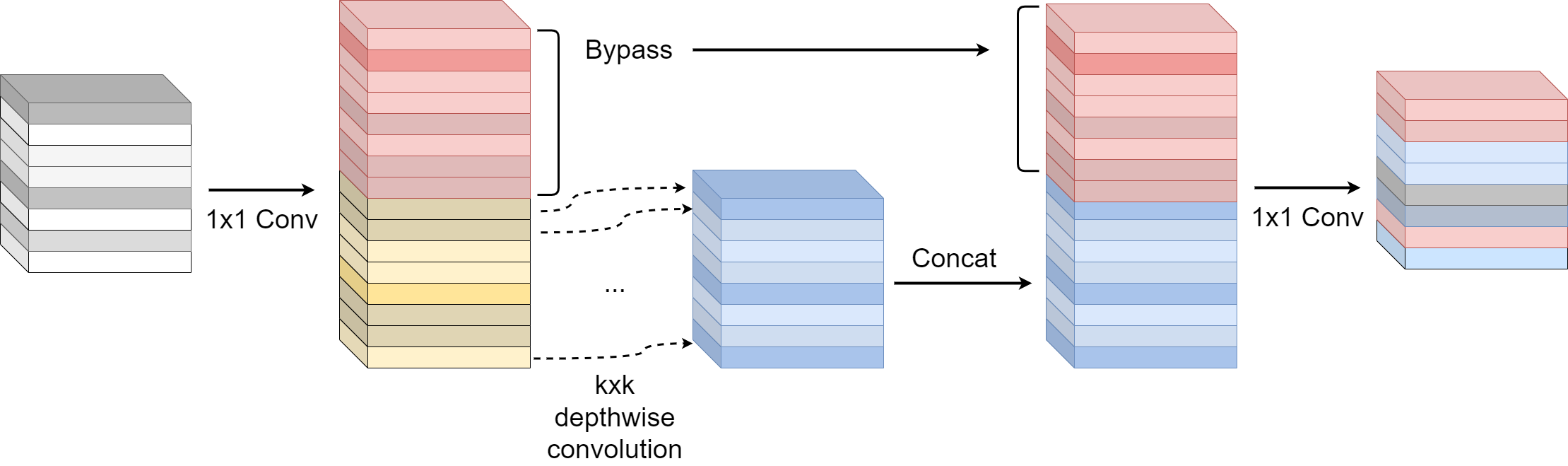}
  \captionof{figure}{Our proposed architecture, inspired by Inverted Bottlenecks \cite{howard_mobilenets_2017, howard_searching_2019} and GhostNets \cite{han_ghostnet_2020, chen_repghost_2022}}
  \label{fig:nodepth}
\end{minipage}%
\begin{minipage}{.49\textwidth}
  \centering
  \includegraphics[width=.7\linewidth]{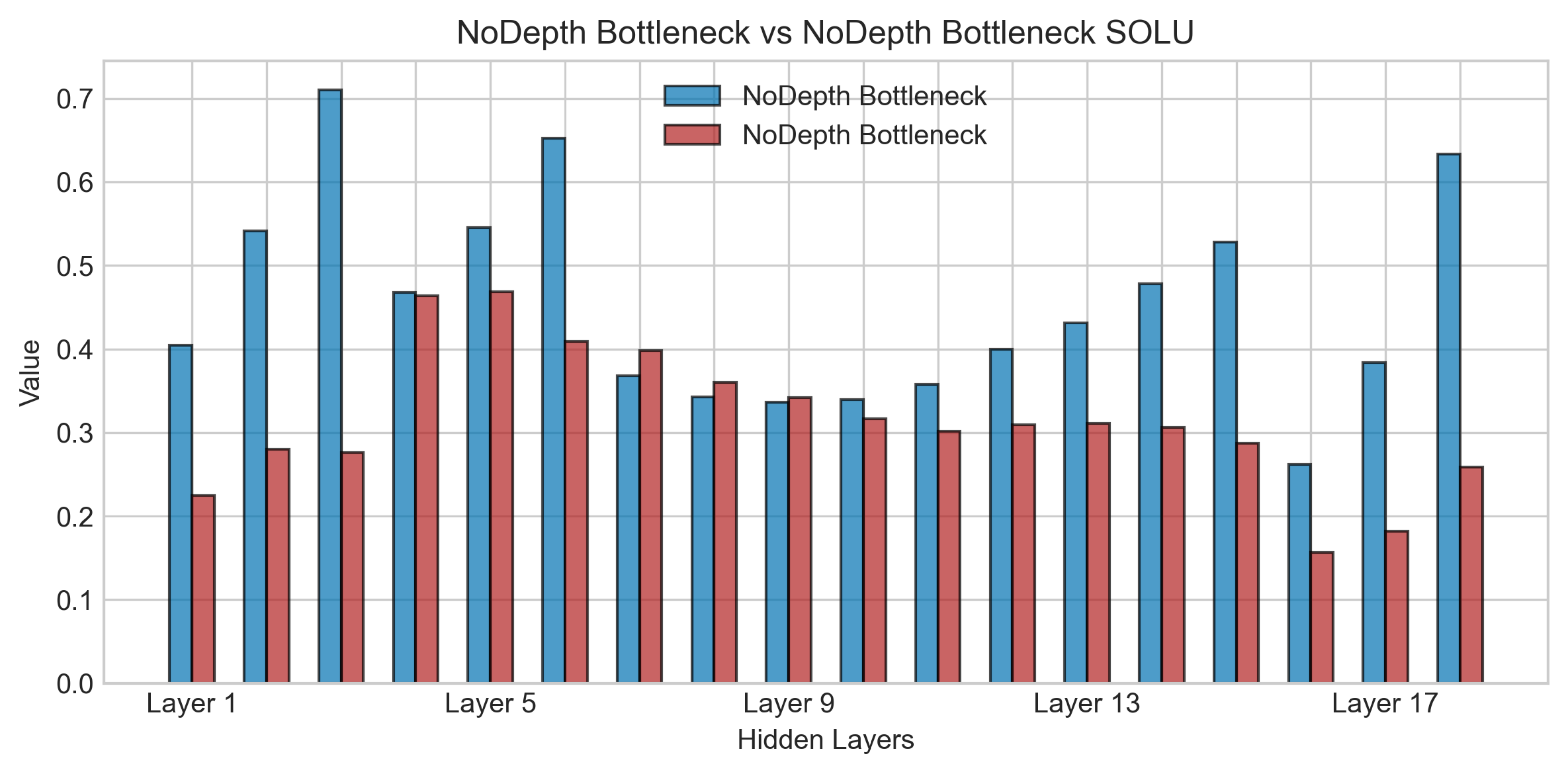}
  \captionof{figure}{NoDepth Bottleneck features alignment, which follows the same behavior using its base architecture and the SoLU one.}
  \label{fig:nodepth_distribution}
\end{minipage}
\end{figure}

To validate our previous experiments, we design a model that combines the best insights uncovered throughout this paper. While the module is computationally expensive due to multiple permutations and concatenations, it primarily serves as a proof of concept for efficient low-parameter scaling.
Our approach confirms key findings, starting with \textbf{depthwise convolution operating at the same dimension as the input residual stream}. To further support this, we position depthwise convolution within the Inverted Bottleneck—the same structure that struggled with the SoLU module. We also replace Batch Norm with Layer Norm to minimize the performance gap between the base and SoLU implementations.


We use a bypass mechanism that retains all dimensions from the first expansion (Figure \ref{fig:nodepth}). Figure \ref{fig:nodepth_distribution} highlights the bottleneck mechanism we want to emphasize in this paper: Feature alignment that remains roughly the same despite interference noise reduction.
To prove its theoretical scaling superiority, following MobileNetv1's low-parameter scaling, we compare the proof-of-concept on a more complex dataset: ImageNet.

\begin{figure}
\centering
\includegraphics[width=0.5\linewidth]{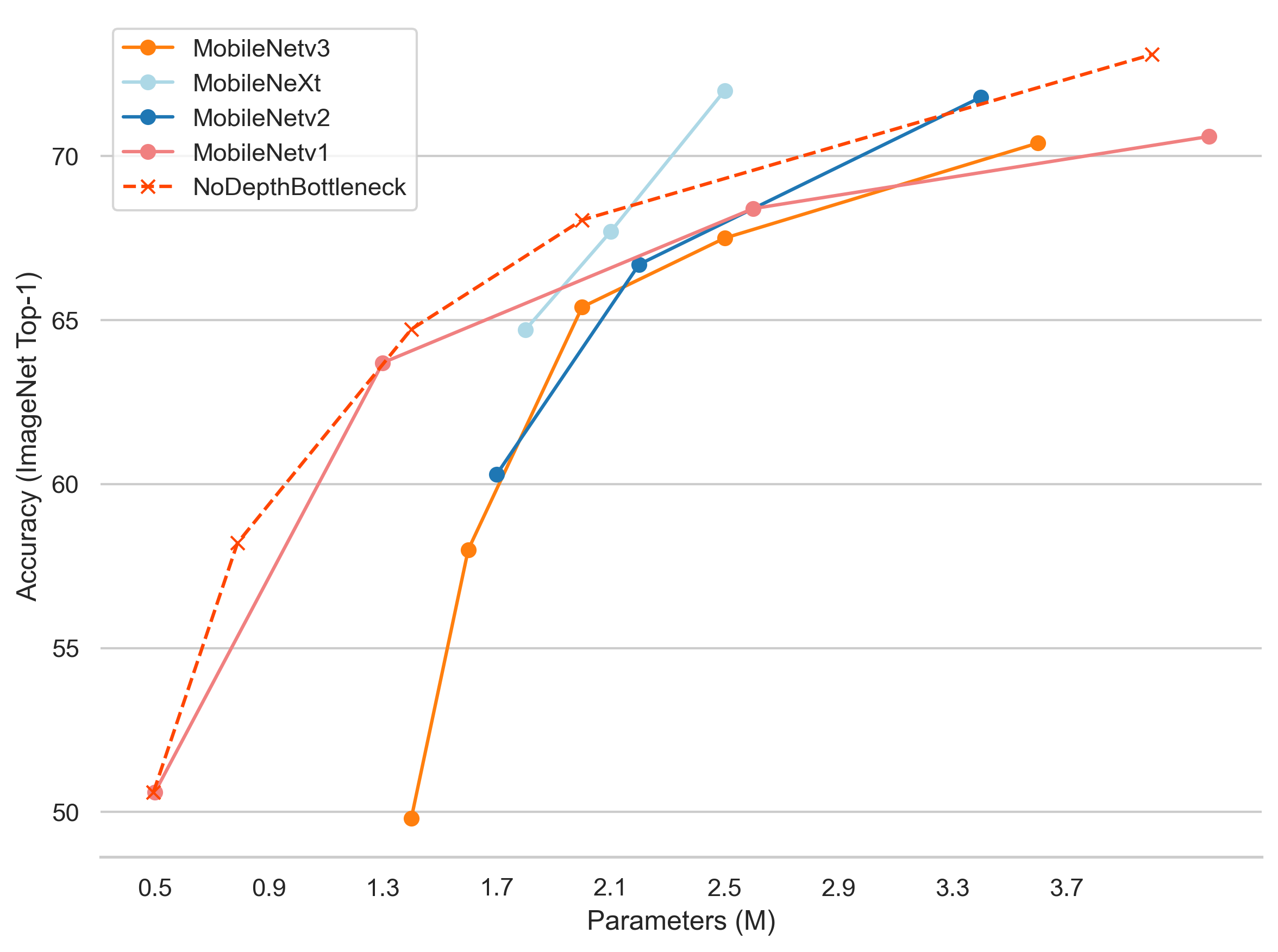}
\captionof{figure}{NoDepth Bottleneck with sota bottleneck architectures}
\label{fig:imagenetsota}
\end{figure}

On ImageNet, our proof of concept enhances the scaling of MobileNetv1, maintaining similar performance in ultra-low-parameter scenarios while achieving better scaling at higher parameter counts, outperforming MobileNetv2 and v3 (Figure \ref{fig:imagenetsota}).

\section{Conclusion}
\label{sec:conclusion}
Our study examined the mechanistic behavior of bottleneck architectures and the impact of superlinear activation functions, specifically the SoLU module, focusing on low-parameter models for computer vision. Our goal was to understand how these networks manage superposition.

Our key findings highlight several important aspects of interference in low-parameter neural networks. First, we examined how interference affects performance and identified its limitations. Our experiments show that MobileNeXt (with Batch Norm), MobileNetv2, and MobileNetv3 struggle to scale efficiently due to their reliance on interference in critical architectural components, which proves ineffective at very low scales.

Next, we explored the relationship between superlinear functions and neural network architecture. Our results indicate that specific models achieve better scaling by reducing their dependence on interference, especially at low parameter counts. We identified architectures such as MobileNetv1, ConvNeXt, MobileNeXt (Layer Norm), and NoDepth Bottleneck as less affected by SoLU, offering insights into interference-free training.

Finally, we analyzed feature alignment by studying feature map vector directions. Models most impacted by SoLU tended to generate feature maps where multiple features align within dissimilar high-dimensional spaces. 

Our findings suggest that reducing interference can improve low-scale model performance in computer vision. We demonstrated why MobileNets and architectures incorporating components like ConvNeXt achieve better scalability by optimizing architectures to minimize reliance on superposition. Limiting interference dependence brings advancements in embedded AI and optimization for resource-constrained devices.

While our study focused on explaining the mechanisms behind widely used bottleneck architectures, it did not introduce a new state-of-the-art design. Future research should investigate whether improving superposition within interference-free architectures leads to better results. 

\section*{Acknowledgement}
This work was supported by Chips Joint Undertaking (Chips JU) in EdgeAI “Edge AI Technologies for Optimised Performance Embedded Processing” project, grant agreement No 101097300 and EdgeAI-Trust «Decentralized Edge Intelligence: Advancing Trust, Safety, and Sustainability in Europe» project has received funding from Chips Joint Undertaking (Chips JU) under grant agreement No 101139892.

\newpage

\printbibliography[heading=subbibintoc]

\end{document}